# Lazy Evaluation of Symmetric Bayesian Decision Problems

**Anders L. Madsen**    **Finn V. Jensen**
Department of Computer Science
Aalborg University
Fredrik Bajers Vej 7C, DK-9220 Aalborg, Denmark

## Abstract

Solving symmetric Bayesian decision problems is a computationally intensive task to perform regardless of the algorithm used. In this paper we propose a method for improving the efficiency of algorithms for solving Bayesian decision problems. The method is based on the principle of lazy evaluation - a principle recently shown to improve the efficiency of inference in Bayesian networks. The basic idea is to maintain decompositions of potentials and to postpone computations for as long as possible. The efficiency improvements obtained with the lazy evaluation based method is emphasized through examples. Finally, the lazy evaluation based method is compared with the HUGIN and valuation-based systems architectures for solving symmetric Bayesian decision problems.

## 1 INTRODUCTION

Bayesian decision theory provides a solid foundation for assessing and thinking about actions under uncertainty. A symmetric Bayesian decision problem is specified with a set of decision variables, a set of chance variables, a multiplicative decomposition of the joint probability distribution of the chance variables given the decision variables, and a utility function specifying the preferences of the decision maker. Solving Bayesian decision problems is, unfortunately, a computationally intensive task to perform.

Influence diagrams [Howard and Matheson, 1984] is an effective modeling framework for analysis of symmetric Bayesian decision making under uncertainty. The influence diagram is a natural representation for capturing the semantics of decision making with a minimum of clutter and confusion for the decision maker [Shachter and Peot, 1992]. An influence diagram is essentially a Bayesian network augmented with decision variables and utility nodes.

Valuation-based systems (VBS) [Shenoy, 1992] is a framework different from influence diagrams for solving symmetric Bayesian decision problems. Influence diagrams are based on the semantics of conditional independence relations while VBS are based on a factorization of the joint probability distribution of the chance variables and a decomposition of the utility function. A graphical illustration of a VBS is referred to as a valuation network. In valuation networks variables, potentials and the precedence constraints are specified explicitly.

Solving a symmetric Bayesian decision problem amounts to computing an optimal strategy maximizing the expected utility for the decision maker and computing the maximal expected utility of this strategy. An optimal strategy specifies the optimal decision for each decision variable given the past. A number of different methods for solving symmetric Bayesian decision problems exists, see for instance [Shachter, 1986, Shenoy, 1992, Ndilikilikesha, 1994, Jensen et al., 1994, Zhang, 1998].

Recently, the lazy propagation architecture for probabilistic inference in Bayesian networks was proposed [Madsen and Jensen, 1998]. The method we propose is based on applying the principles of lazy evaluation to solving decision problems. Bayesian network inference involves elimination of chance variables only. Solving symmetric Bayesian decision problems, on the other hand, involves elimination of both chance and decision variables. Elimination of a chance variable is performed by summation while elimination of a decision variable is performed by maximization. Extending the lazy propagation architecture for solving decision problems involves extending both the on-line triangulation heuristic, the elimination algorithm, and the potential combination algorithm.

In general, we assume the reader to be familiar with Bayesian networks, influence diagrams, and VBS.

## 2  PRELIMINARIES

The set of chance variables of a symmetric Bayesian decision problem is partitioned into disjoint information sets $\mathcal{I}_0, \mathcal{I}_1, \ldots, \mathcal{I}_n$ relative to the decision variables. The partition induces a partial ordering $\prec$ on the variables of the decision problem. The set of variables observed between decision $D_i$ and $D_{i+1}$ precedes $D_{i+1}$ and succeeds $D_i$ in the ordering: $\mathcal{I}_0 \prec D_1 \prec \mathcal{I}_1 \cdots \prec D_n \prec \mathcal{I}_n$.

The set of legal elimination orderings is constrained to include only orderings where all variables in information set $I_i$ are eliminated before decision $D_i$. This implies that $I_0$ is the set of chance variables initially observed and $I_n$ is the set of chance variables never observed or observed after the last decision has been made. A total ordering on the decision variables is usually assumed. This assumption can, however, be relaxed. [Nielsen and Jensen, 1999] describes when decision problems with only a partial ordering on the decision variables is well-defined.

One of the main difficulties in solving symmetric Bayesian decision problems is that elimination of chance and decision variables do not, in general, commute. Variables within the same information set can, however, be eliminated in any order and consecutive decision variables $D_i$ and $D_{i+1}$ can be eliminated in any order if the information set $\mathcal{I}_i$ is empty.

A decision rule for decision variable $D_i$ is a function $\delta_i : \text{dom}(\pi_{D_i}) \to \text{dom}(D_i)$ where $\pi_{D_i}$ is the relevant past of $D_i$. The relevant past of decision variable $D_i$ is the subset of the informational parents of $D_i$ relevant for making decision $D_i$ (see [Nielsen and Jensen, 1999] for a structural definition).

A strategy is an ordered set of decision rules $\Delta = (\delta_1, \ldots, \delta_n)$ including a decision rule for each decision. An optimal strategy $\hat{\Delta}$ returns the optimal decision for the decision maker to make at each decision. To solve a decision problem is to compute an optimal strategy $\hat{\Delta}$ maximizing the expected utility for the decision maker and to compute the maximum expected utility $\text{MEU}(\hat{\Delta})$ of $\hat{\Delta}$.

The utility function specifying the preferences of the decision maker is most often assumed to decompose additively as when the utility function decomposes multiplicatively solving the decision problem reduces to a task similar to performing probabilistic inference in Bayesian networks. We return to this point later.

The perfect recall assumption of Bayesian decision problems implies that at the time of any decision, the decision maker remembers all past decisions and all previously known information. This implies that a decision variable and all its parents are informational parents of all subsequent decision variables (non-forgetting arcs will not be depicted in any figure, but they will always be assumed present).

For ease of exposition and limited space we focus on the influence diagram representation of symmetric Bayesian decision problems. An influence diagram ID is a triple $(G, P, U)$ where $G$ is a directed acyclic graph with chance, decision, and utility nodes. Each chance node corresponds to a chance variable which is associated with a conditional probability distribution. The set of conditional probability distributions is denoted $P$. Each value node is associated with a utility potential and the set of utility potentials is denoted $U$. Each decision node is associated with a decision variable. Arcs into chance nodes denote probabilistic dependence relations and an arc from a node $X$ into a decision node $D$ indicates that the state of $X$ is known when decision $D$ is to be made.

In the rest of the paper the generalized marginalization operator $\mathsf{M}$ introduced by [Jensen et al., 1994] is used. The marginalization operator works differently for marginalization of chance and decision variables:

$$\mathsf{M}_X \rho = \sum_X \rho \quad \text{and} \quad \mathsf{M}_D \rho = \max_D \rho,$$

where $X$ is a chance and $D$ is a decision variable.

## 3  THE HUGIN ARCHITECTURE

The HUGIN architecture for solving influence diagrams is based on message passing in strong junction trees. A strong junction tree representation $T$ of an influence diagram ID is constructed based on moralization, removal of the information arcs, and removal of the direction on the remaining arcs of ID. Let $\text{ID}^m$ denote the resulting graph. The nodes of $T$ are cliques of a strong triangulation of $\text{ID}^m$ and $T$ has the property that for any two adjacent cliques $C_i$ and $C_j$ with $C_i$ closest to the strong root $R$ of $T$ there exists an ordering of the variables of $C_j$ respecting $\prec$ with the variables of $C_i \cap C_j$ preceding the variables of $C_j \setminus C_i$.

Once $T$ is constructed, the probability and utility potentials of ID are associated with cliques of $T$ such that the domain of the potential is a subset of the clique domain. When all potentials have been associated with cliques, the probability potentials associated with a clique $C$ are combined by multiplication and the utility potentials associated with $C$ are combined by summation to form the initial clique probability and utility potentials $\phi_C$ and $\psi_C$, respectively.

Messages are passed from the leaf cliques of T to R and messages in different branches of T can be passed independently. The message passed over a separator S connecting cliques $C_i$ and $C_j$ consists of two potentials, a probability potential $\phi_S^*$ computed by marginalization of $\phi_{C_j}$ down to S and a utility potential $\psi_S^*$ computed by marginalization of the combination of $\phi_{C_j}$ and $\psi_{C_j}$ down to S:

$$\phi_S^* = \underset{C_j \setminus S}{\mathsf{M}} \phi_{C_j} \quad \text{and} \quad \psi_S^* = \underset{C_j \setminus S}{\mathsf{M}} \phi_{C_j} \psi_{C_j}.$$

When a clique C receives messages from neighboring separators the clique probability and utility potentials $\phi_C$ and $\psi_C$ are updated with the potentials received:

$$\phi_C^* = \phi_C \prod_{S \in ch(C)} \phi_S^* \quad \text{and} \quad \psi_C^* = \psi_C + \sum_{S \in ch(C)} \frac{\psi_S^*}{\phi_S^*},$$

where $ch(C)$ is the set of separators between C and all adjacent cliques further away from R.

The relevant past of a decision variable is not simple to read off from the strong junction tree structure. However, let C be the clique containing decision variables D closest to R and let $V_D$ be the set of variables preceding D in C. If and only if $X \in V_D$, then there is a path between X and D in $ID^m$ such that when D is to be eliminated all variables on the path between X and D have already been eliminated. This implies that all past variables relevant for computing the optimal decision rule $\delta$ is a subset of $V_D$ where $\delta$ returns the maximizing alternatives of the utility potential $\psi(D, V_D)$.

The partial ordering imposed by the structure of the influence diagram ID is extended to a total ordering by the heuristic method used for the strong triangulation of $ID^m$. The strong junction tree T constructed from the strong triangulation imposes a partial order on the elimination of variables during message passing. The fact that T only imposes a partial order on the set of legal elimination orderings cannot be exploited fully in the HUGIN architecture as the potentials initially associated with cliques are combined to form initial clique probability and utility potentials.

## 4 VARIABLE ELIMINATION

The maximum expected utility of a symmetric Bayesian decision problem is calculated by eliminating all variables of the decision problem in the reverse order of the partial ordering imposed by the information constraints. A formula representing the task of computing $\mathsf{MEU}(\hat{\Delta})$ by eliminating a set of variables V consisting of chance variables $\{X_1, \ldots, X_n\}$ and decision variables $\{D_1, \ldots, D_o\}$ from a set of probability potentials $\Phi = \{P(X_i | pa(X_i)) | 1 \leq i \leq n\}$ and a set of utility potentials $\Psi = \{U_1, \ldots, U_m\}$ is:

$$\mathsf{MEU}(\hat{\Delta}) = \underset{X \in V}{\mathsf{M}} \left( \left( \prod_{i=1}^n P(X_i | pa(X_i)) \right) \sum_{j=1}^m U_j \right). \quad (1)$$

Assume the first variable to eliminate according to the strong elimination order is Y. The set of utility potentials $\Psi$ can be divided into two subsets. Let $\mathcal{D}_Y$ be the subset of utility potentials including Y in the domain: $\mathcal{D}_Y = \{U_j | Y \in dom(U_j), 1 \leq j \leq, m\}$, let $\mathcal{N}_Y$ be the set of utility potentials not including Y in the domain: $\mathcal{N}_Y = \{U_j | 1 \leq j \leq m\} \setminus \mathcal{D}_Y$, and let $\mathcal{H}_Y = Y \cup ch(Y)$ where $ch(Y)$ is the set of children of Y. Furthermore, let $\phi_Y$ be the potential obtained by eliminating Y from the combination of all probability potentials with Y in the domain and let $\psi_Y$ be the potential obtained by eliminating Y from the combination of all probability and utility potentials with Y in the domain:

$$\phi_Y = \underset{Y}{\mathsf{M}} \prod_{X_i \in \mathcal{H}_Y} P(X_i | pa(X_i)),$$

$$\psi_Y = \underset{Y}{\mathsf{M}} \left( \prod_{X_i \in \mathcal{H}_Y} P(X_i | pa(X_i)) \right) \sum_{U_k \in \mathcal{D}_Y} U_k. \quad (2)$$

With these definitions, equation 1 is rewritten as:

$$\mathsf{MEU}(\hat{\Delta}) = \underset{X \in V}{\mathsf{M}} \left[ \left( \prod_{i=1}^n P(X_i | pa(X_i)) \right) \right.$$
$$\left. \left( \sum_{U_j \in \mathcal{N}_Y} U_j + \sum_{U_k \in \mathcal{D}_Y} U_k \right) \right]$$
$$= \underset{X \in V \setminus \{Y\}}{\mathsf{M}} \left[ \left( \prod_{X_j \in V \setminus \mathcal{H}_Y} P(X_j | pa(X_j)) \right) \right.$$
$$\underset{Y}{\mathsf{M}} \left( \prod_{X_j \in \mathcal{H}_Y} P(X_j | pa(X_j)) \right) \left( \sum_{U_j \in \mathcal{N}_Y} U_j + \sum_{U_k \in \mathcal{D}_Y} U_k \right) \right]$$
$$= \underset{X \in V \setminus \{Y\}}{\mathsf{M}} \left[ \left( \prod_{X_j \in V \setminus \mathcal{H}_Y} P(X_j | pa(X_j)) \right) \right.$$
$$\left. \left( \left( \sum_{U_j \in \mathcal{N}_Y} U_j \right) \phi_Y + \psi_Y \right) \right] \quad (3)$$
$$= \underset{X \in V \setminus \{Y\}}{\mathsf{M}} \left[ \left( \prod_{X_j \in V \setminus \mathcal{H}_Y} P(X_j | pa(X_j)) \right) \phi_Y \right.$$
$$\left. \left( \sum_{U_j \in \mathcal{N}_Y} U_j + \frac{\psi_Y}{\phi_Y} \right) \right]. \quad (4)$$

Equation 4 shows that elimination of Y from $\Phi$ and $\Psi$ produces a probability potential $\phi_Y$ and a utility potential $\frac{\psi_Y}{\phi_Y}$. When Y is eliminated, all potentials of $\Phi_Y$ and $\Psi_Y$ including Y in their domain are removed from $\Phi$ and $\Psi$, respectively. Instead, $\phi_Y$ and $\frac{\psi_Y}{\phi_Y}$ are added. The union of the updated sets $\Phi^* \cup \Psi^*$ is a value preserving reduction of $\Phi \cup \Psi$ where Y has been eliminated.

The above derivation is similar to the derivation presented in [Dechter, 1996] with the exception that we do not assume the decision variables to be root variables in the influence diagram. Equation 4 is essentially a general specification of the fusion operation of [Shenoy, 1992]. The fusion operation, however, explicitly differentiates between four different cases depending on the variable Y being eliminated and the set of potentials with Y in the domain.

One of the main difficulties in solving symmetric Bayesian decision problems is that an additive decomposition of the utility function makes combination of utility and probability potentials non-associative. If the utility function decomposes multiplicatively, then combination of probability and utility potentials is associative, and the need for division and summation of potentials is eliminated. This implies that the task of solving a symmetric Bayesian decision problem is reduced to a task similar to performing probabilistic inference in a Bayesian network [1].

## 5 LAZY PROPAGATION

For ease of exposition the lazy evaluation based method is described in the context of influence diagrams and strong junction trees, but notice that the principles of lazy evaluation can be applied to other representations and computational structures such as VBS. In the context of strong junction trees we refer to the method as the lazy propagation algorithm.

Lazy propagation is based on message passing in a strong junction tree representation, say T, of a decision problem represented as an influence diagram, say ID. Initially, the probability and utility potentials of ID are associated with cliques of T. The potentials associated with cliques are not combined, so during message passing each clique and separator holds two sets of potentials. Messages are passed from the leaves of T towards the strong root R by recursively invoking the *CollectMessage* algorithm (defined below). When *CollectMessage* is invoked on a clique $C_j$ from an adjacent clique $C_i$, clique $C_j$ invokes *CollectMessage* on all other adjacent cliques. When these cliques have fin-

---
[1] A multiplicative decomposition of the utility function is obtainable using the exponential function, for instance.

ished their *CollectMessage*, $C_j$ absorbs messages from each of them and $C_i$ absorbs from $C_j$:

**Algorithm 5.1 [CollectMessage]**
Let $C_i$ and $C_j$ be adjacent cliques. If *CollectMessage* is invoked on $C_j$ from $C_i$, then:

1. $C_j$ invokes *CollectMessage* on all adjacent cliques except $C_i$.

2. The message from $C_j$ to $C_i$ is absorbed (algorithm 5.2).

□

From the description of the *CollectMessage* algorithm it is clear that information is passed between adjacent cliques by absorption. Consider two adjacent cliques $C_i$ and $C_j$ separated by S. Absorption from $C_j$ to $C_i$ over S amounts to eliminating the variables of $C_j \setminus S$ from the sets of probability and utility potentials $\Phi$ and $\Psi$ associated with $C_j$ and the separators of $ch(C_j)$ and then associating the obtained potentials with S:

**Algorithm 5.2 [Absorption]**
Let $C_i$ and $C_j$ be adjacent cliques and let S be the separator between $C_i$ and $C_j$. If *Absorption* is invoked on $C_j$ from $C_i$, then:

1. Set $\mathcal{R}_S = \Phi_{C_j} \cup \Psi_{C_j} \cup \bigcup_{S' \in ch(C_j)} \Phi^*_{S'} \cup \Psi^*_{S'}$.

2. Marginalize out all variables $\{Y \in dom(\rho) \mid \rho \in \mathcal{R}_S, Y \notin S\}$ (algorithm 5.3). Let $\Phi^*_S$ and $\Psi^*_S$ be the sets of probability and utility potentials obtained.

3. Associate $\Phi^*_S$ and $\Psi^*_S$ with S as the sets of potentials passed from $C_j$ to $C_i$.

□

Algorithm 5.2 uses the algorithm for marginalization of variables from two sets of potentials:

**Algorithm 5.3 [Marginalization]**
Let $\Phi$ and $\Psi$ be sets of probability and utility potentials, respectively. If *Marginalization* of variable Y is invoked on $\Phi \cup \Psi$, then:

1. Set $\Phi_Y = \{\phi \in \Phi \mid Y \in dom(\phi)\}$ and $\Psi_Y = \{\psi \in \Psi \mid Y \in dom(\psi)\}$.

2. Calculate

$$\phi^*_Y = \underset{Y}{\mathsf{M}} \prod_{\phi \in \Phi_Y} \phi,$$

$$\psi^*_Y = \underset{Y}{\mathsf{M}} \prod_{\phi \in \Phi_Y} \phi \sum_{\psi \in \Psi_Y} \psi,$$

3. Return $\Phi^* = \Phi \cup \{\phi_Y^*\} \setminus \Phi_Y$ and $\Psi^* = \Psi \cup \left\{\frac{\psi_Y^*}{\phi_Y^*}\right\} \setminus \Psi_Y$.

□

The maximizing alternatives of the utility potential $\psi(D, V_D)$ from which D is eliminated during evaluation of the influence diagram should be recorded as the optimal decision rule for D.

**Theorem 5.1 [Lazy Propagation]**
*Lazy propagation computes the optimal strategy $\hat{\Delta}$ maximizing the expected utility for the decision maker and the maximum expected utility* $MEU(\hat{\Delta})$.

*Proof.* Nothing to prove. It is just a change of representation. □

### 5.1 EFFICIENCY IMPROVEMENTS

A number of adjustments to the basic lazy propagation algorithm can improve the efficiency of the algorithm considerably. Consider absorption of information between adjacent cliques $C_i$ and $C_j$. In step 1 of algorithm 5.2 all potentials associated with clique $C_j$ and separators of $ch(C_j)$, $\Phi$ and $\Psi$ say, are assumed relevant for calculating the message to pass to clique $C_i$. The subset of potentials $\mathcal{R}_S$ that can be relevant for computing the message to pass over S can be determined in time linear in the size of the domain graph induced by $\Phi \cup \Psi$ using the d-separation criterion [Geiger et al., 1990]:

**Algorithm 5.4 [Find Potentials That Can Be Relevant]**
Let $\Phi$ and $\Psi$ be sets of probability and utility potentials and let S be a set of variables. If *Find Potentials That Can Be Relevant* for calculating the joint of S is invoked on $\Phi \cup \Psi$, then:

1. Return $\mathcal{R}_S = \{\rho \in \Phi \cup \Psi \mid \exists X \in dom(\rho) \text{ and } X \text{ d-connected to } Y \in S\}$.

□

Barren variables in Bayesian networks are defined as variables which are neither evidence nor target variables and have only barren descendants. We define barren variables in influence diagrams to have the same properties as barren variables in Bayesian networks with the additional property that a variable with a directed path to a utility potential cannot be barren. If a variable, X say, is barren when only the set of probability potentials are considered, we refer to X as a *probabilistic barren variable*. If a decision variable D is barren, then any alternative for D is optimal. The set of potentials $\mathcal{R}_S$ returned by algorithm 5.4 can be reduced by recursively removing all potentials including only barren variables as head variables.

If a variable X is probabilistic barren, then $\phi = \mathsf{M}_X \prod_{\phi \in \Phi_X} \phi$ is a unity-potential (referred to as vacuous valuations in [Shenoy, 1992]). During calculation of messages unity-potentials are not calculated and if the denominator of the division introduced in step 3 of algorithm 5.3 is a unity-potential, then the division is not performed. If the denominator is not a unity-potential, then the division should be performed on the domain of $\prod_{\phi \in \Phi_Y} \phi$ and not on the domain of $\psi_Y^*$ as the domain size of $\psi_Y^*$ is at least the size of the domain of $\prod_{\phi \in \Phi_Y} \phi$. A division should not be performed when it is introduced, but should be postponed for as long as possible.

Step 3 of algorithm 5.3 corresponds to equation 4, but in some cases it is more efficient to use equation 3. If, for example, the set $\mathcal{N}_Y$ is known to be empty, then it is more efficient to use equation 3. Whether it is more efficient to use equation 3 or equation 4 cannot always be determined locally in each clique.

Consider the calculation of $\psi_Y$ in equation 2. This calculation can also be performed as:

$$\psi_Y = \mathsf{M}_Y \sum_{U_k \in \mathcal{D}_Y} \left( \prod_{X_i \in \mathcal{H}_Y} P(X_i | pa(X_i)) \right) U_k. \quad (5)$$

Whether to distribute the combination of all probability potentials depends on the structure of the utility potentials and whether or not the division in step 3 of algorithm 5.3 is necessary. If all utility potentials have the same domain, then it is more efficient not to distribute. If, on the other hand, the utility potentials only share the variable Y, then it is more efficient to distribute.

Due to the partial ordering of the variables imposed by the information constraints, a variable $X \in \mathcal{I}_k$ cannot be instantiated by evidence when calculating the maximum expected utility of a decision variable $D_j$ where $k \geq j$. This implies that the probability potential $\phi_X$ calculated when eliminating X is not a unity-potential only if X is non-descendent of $D_j$. In general, evidence instantiating a variable X is exploited by reducing the domain of potentials including X in the domain.

Subject to the limitations of the information constraints, the internal elimination order in a clique C is determined based on the set of potentials associated with C and separators of $ch(C)$. Let V be the set of variables to eliminate when calculating a message. If a variable $X \in V$ is a probabilistic barren variable, then X is eliminated first unless elimination of another non-probabilistic barren variable $Y \in V$ reduces the domain size of a utility potential and the elimination of Y does not make a barren variable non-barren.

## 5.2 PROPERTIES

In this section some properties of the lazy propagation architecture are exemplified.

**Example 5.2**
The influence diagram ID of [Jensen et al., 1994] and a corresponding junction tree T are shown in figures 1 and 2, respectively. Figure 3 shows the flow of messages in T during *CollectMessage* invoked on $BD_1EFD$ (sets including only unity-potentials are not shown).

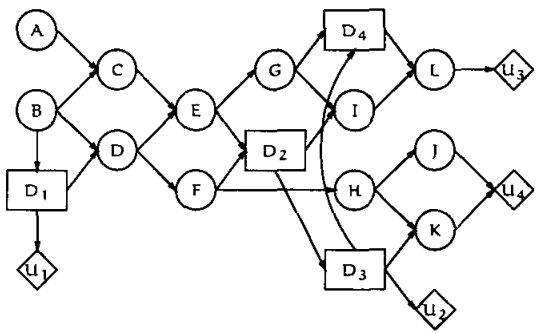

Figure 1: An influence diagram ID.

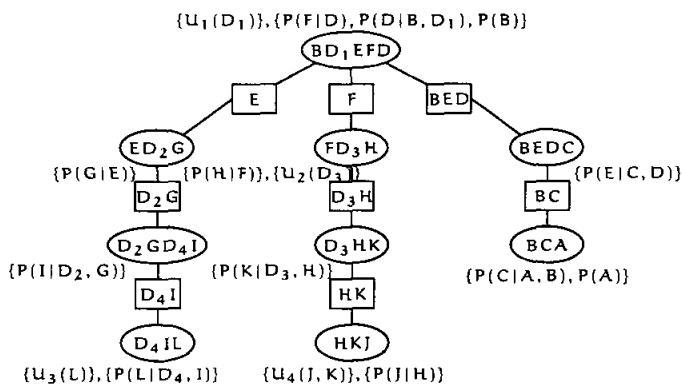

Figure 2: A junction tree for ID shown in figure 1.

Consider the message $\{\psi(D_4, I)\}$ passed from $D_4IL$ to $D_2GD_4I$. This message is calculated as:

$$\phi(D_4, I) = \sum_L P(L|D_4, I) = 1_{D_4, I},$$

$$\psi(D_4, I) = \sum_L U_3(L)P(L|D_4, I).$$

As $\phi(D_4, I)$ is a unity-potential it is not computed and the division introduced in step 3 of algorithm 5.3 is not performed. In fact, no divisions are performed when solving ID in the lazy propagation architecture. Furthermore, as a benefit of the decomposition of potentials and exploitation of unity-potentials considerably fewer arithmetic operations are performed in the lazy propagation architecture than in the HUGIN architecture. If we assume all variables to be binary, then 575 and 161 arithmetic operations are performed in the HUGIN and the lazy propagation architectures, respectively. This does not include the operations required to compute the initial clique probability and utility potentials in the HUGIN architecture.

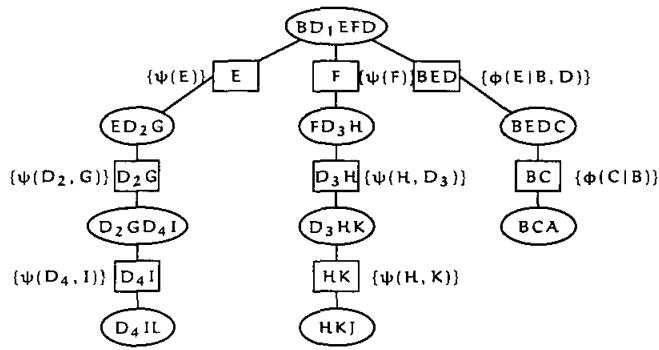

Figure 3: The flow of messages towards $BD_1EFD$.

□

Using the HUGIN architecture it is difficult to handle variables which the decision maker might or might not have evidence on. It is, in fact, necessary to create a strong junction for each scenario. The lazy propagation architecture supports certain types of changes in the structure of the influence diagram without requiring construction of a new strong junction tree. The types of changes supported follows from the way in which variables are eliminated. Variables are eliminated one at a time according to the reverse partial order imposed by the structure of the strong junction tree. If a change made to the structure of the influence diagram moves a variable X from one information set $\mathcal{I}_i$ to another information set $\mathcal{I}_j$ where $j < i$, then X has to be eliminated at a later time than specified in the original influence diagram. This is readily handled in the lazy propagation architecture as this just amounts to postponing the elimination of X until information set $\mathcal{I}_j$ is reached during message passing in the junction tree. The opposite situation, however, cannot be handled as easily. To facilitate the possibility of changing the time of elimination of a variable X the strong junction tree can be constructed under the assumption that $X \in \mathcal{I}_n$.

**Example 5.3 [Myopic Value of Information]**
[Dittmer and Jensen, 1997] introduces a method for myopic value of information analysis (VOI) on a predetermined set of variables V in influence diagrams. The method is based on introducing additional control structures. The control structures can be interpreted as adding V to all cliques of the junction tree, but due to the introduction of additional control struc-tures the decrease in performance is not as serious as this. Similarly to the above described approach, clique and separator potentials are only extended as needed. The method facilitates VOI for different information scenarios within the same junction tree.

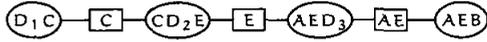

Figure 4: The original junction tree T.

Consider the junction tree T with root $D_1C$ shown in figure 4 (borrowed from [Dittmer and Jensen, 1997]). To support VOI on B control structures corresponding to extending all cliques and separators of T with B are added to T, see figure 5. The additional control structures facilitate VOI where the time of elimination of B can be chosen arbitrarily.

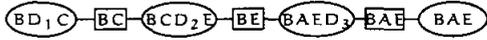

Figure 5: The junction tree T' supporting VOI on B.

Performing VOI on B in the lazy propagation architecture is based on T. By postponing the elimination of B as explained above it is possible to choose the time of elimination of B arbitrarily. □

Much effort has been put into developing various methods for reducing the complexity of Bayesian network inference. Methods for approximation, exploitation of independence of causal influence, and exploitation of context specific independence, for instance, exists for Bayesian networks. Very little effort, however, has been put into developing similar methods for influence diagrams. The next example shows that independence of causal influence (ICI) is readily exploited in the lazy propagation architecture for solving influence diagrams.

**Example 5.4 [Independence of Causal Influence]** Consider the influence diagram ID and the corresponding strong junction tree T shown in figure 6. Assume that $D_1$, $C_2$, and $C_3$ are causally independent with respect to $E_1$ and that all variables are binary with states f and t. In order to compute $MEU(\hat{\Delta})$ all variables have to be eliminated according to the partial order $D_1 \prec \{E_1\} \prec D_2 \prec \{E_2, C_2, C_3\}$.

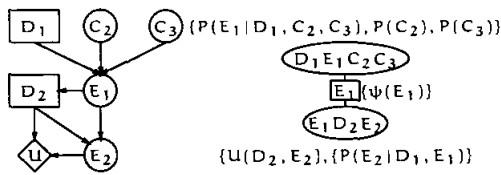

Figure 6: A simple influence diagram and a corresponding strong junction tree T.

If $\psi(E_1)$ is the message passed from $E_1D_2E_2$ to $D_1E_1C_2C_3$, then $MEU(\hat{\Delta})$ is calculated as:

$$MEU(\hat{\Delta}) = \max_{D_1} \sum_{E_1} \psi(E_1) \sum_{C_2} P(C_2) \sum_{C_3} P(C_3)P(E_1|D_1,C_2,C_3).$$

By exploiting ICI the complexity of calculating $MEU(\hat{\Delta})$ can be reduced to linear in the number of parents of $E_1$. Assume that the lazy parent divorcing method developed for Bayesian networks [Madsen and D'Ambrosio, 1999] is used to exploit ICI. One hidden variable Y is introduced as a child of $C_2$ and $C_3$ ($P(E_1^C|C)$ is the contribution from $C \in \{D_1, C_2, C_3\}$ to $E_1$ when $\forall_{C \neq C'} C' = f$):

$$MEU(\hat{\Delta}) = \max_{D_1} \sum_{E_1} \psi(E_1) \sum_{C_2} P(C_2) \sum_{C_3} P(C_3)$$
$$\sum_{\{E_1^{D_1}, Y | E_1 = E_1^{D_1} * Y\}} P(E_1^{D_1}|D_1)$$
$$\sum_{\{E_1^{C_2}, E_1^{C_3} | Y = E_1^{C_2} * E_1^{C_3}\}} P(E_1^{C_2}|C_2)P(E_1^{C_3}|C_3)$$
$$= \max_{D_1} \sum_{E_1} \psi(E_1) \sum_{\{E_1^{D_1}, Y | E_1 = E_1^{D_1} * Y\}} P(E_1^{D_1}|D_1)$$
$$\sum_{\{E_1^{C_2}, E_1^{C_3} | Y = E_1^{C_2} * E_1^{C_3}\}} \sum_{C_2} P(C_2)P(E_1^{C_2}|C_2)$$
$$\sum_{C_3} P(C_3)P(E_1^{C_3}|C_3).$$

When the ICI is exploited, the domain size of the largest potential created when computing $MEU(\hat{\Delta})$ is 3 instead of 4 variables. □

## 6 COMPARISON

The HUGIN architecture for solving symmetric Bayesian decision problems represented as influence diagrams is based on constructing a strong junction tree representation of the influence diagram. The decision problem is solved by message passing in the strong junction tree as explained in section 3. Combining the probability and utility potentials associated with each clique to form the initial clique and utility potentials makes it impossible to benefit from determining the internal elimination order on-line and to take advantage of independence relations induced by evidence. Furthermore, it also makes it impossible to reduce the complexity of inference by exploiting barren and probabilistic barren variables. Barren variables can, however, be determined in a preprocessing step before the strong junction tree is constructed.

[Jensen et al., 1994] argues that the contribution from the division operation plays a smaller rôle in the HUGIN architecture than in VBS since divisions are performed on separators only. We will not discuss this statement in detail here, but only note that a substantial number of the divisions performed in the HUGIN architecture are unnecessary (see example 5.2). The lazy propagation architecture often detects the situations in which the divisions are unnecessary. We believe that it is only rarely necessary to actually perform most of the divisions as indicated in example 5.2. A number of empirical tests or an in depth analysis should be performed to dissolve this. The divisions performed in the lazy propagation architecture are not performed on separators, but are introduced when variables are eliminated. This implies that when a division in fact is performed the operation might have a higher computational cost than in the HUGIN architecture. It is, however, also possible that the converse is the case. That is, because divisions are introduced each time a variable is eliminated and only performed when it cannot be detected to be unnecessary, the operation is likely to play a smaller rôle in the lazy propagation architecture than in the HUGIN architecture.

Recently, a method for solving symmetric Bayesian decision problems represented as influence diagrams by reducing influence diagram evaluation to Bayesian network inference problems has been proposed [Zhang, 1998]. The reduction is performed by repeated use of Cooper's trick and the method induces simpler inference problems than the methods proposed by [Cooper, 1988] and [Shachter and Peot, 1992]. Reducing influence diagram evaluation to Bayesian network inference problems makes it possible and easy to exploit existing methods for improving the efficiency of Bayesian network inference to improve the efficiency of decision problem solving. We have argued that it is unnecessary to convert to Bayesian inference problems in order to exploit methods for improving the efficiency of decision problem solving. Exploitation of ICI in the lazy propagation architecture is just one example.

The VBS architecture for solving decision problems is based on the fusion algorithm. The marginalization algorithm (algorithm 5.3) of the basic lazy propagation architecture is essentially the fusion algorithm of VBS. When the marginalization algorithm is extended with the efficiency improvements of section 5.1 the algorithms become different:

**Example 6.1 [Valuation-based systems]**

Consider the valuation network shown in figure 7 (borrowed from [Shenoy, 1992]). The valuation network specifies a joint probability distribution for $C_1$ and $C_2$ (the triangle in the figure), a utility function decomposing additively into utility potentials $U(C_1)$ and $U(D, C_2)$, and the precedence constraints.

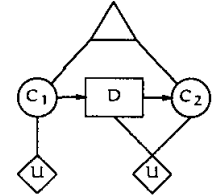

Figure 7: A valuation network for a decision problem.

The computations performed in the VBS corresponds to the following expression:

$$MEU(\hat{\Delta}) = \sum_{C_1} \left( \sum_{C_2} P(C_1, C_2) \right) \\ \left( U(C_1) + \max_D \sum_{C_2} U(D, C_2) \frac{P(C_1, C_2)}{\sum_{C_2} P(C_1, C_2)} \right). \quad (6)$$

The corresponding expression obtained with the lazy evaluation based method is:

$$MEU(\hat{\Delta}) = \sum_{C_1} \left( \sum_{C_2} P(C_1, C_2) \right) \\ \left( U(C_1) + \max_D \left( \sum_{C_2} U(C_2, D) \frac{P(C_1, C_2)}{\sum_{C_2} P(C_1, C_2)} \right) \right) \\ = \sum_{C_1} \left( \sum_{C_2} P(C_1, C_2) \right) \\ \left( U(C_1) + \left( \frac{\max_D \sum_{C_2} P(C_1, C_2) U(C_2, D)}{\sum_{C_2} P(C_1, C_2)} \right) \right) \\ = \sum_{C_1} \left( \sum_{C_2} P(C_1, C_2) U(C_1) + \\ \max_D \sum_{C_2} P(C_1, C_2) U(C_2, D) \right). \quad (7)$$

The division performed in equation 6 is avoided if the division is postponed until it can be determined that it is unnecessary to perform it. The above derivation shows how to avoid the division. The number of arithmetic operations performed to calculate equation 6 is 25 while only 21 are operations performed to compute equation 7 (the four divisions are avoided). □

When described in the context of valuation networks the lazy evaluation based method and the VBS architecture are very similar, but they are as example 6.1 shows different. If we consider the lazy evaluation based method in the context of strong junction trees, the differences between the two methods become clear. The main focus of VBS is on elimination of variables while the main focus of lazy propagation is on message passing. The messages are calculated by elimination of variables and the order in which variables are eliminated is controlled by the structure of the strong junction tree and the algorithms described in section 5.

Lazy propagation exploits the structure of the strong junction tree to simplify tasks such as finding the set of potentials relevant for eliminating a particular variable. The fusion operation, on the other hand, is applied directly to the entire set of potentials of the VBS without any preprocessing.

# 7 CONCLUSION

We have presented a lazy evaluation based method for solving symmetric Bayesian decision problems. The lazy evaluation based method is not developed to a particular computational structure as the principles of the method applies to various different computational structures. We have, however, in detail described how the principles of lazy evaluation can be applied to influence diagrams and strong junction trees. We have only briefly indicated how the principles of lazy evaluation can be applied to VBS. The lazy evaluation principle offers the opportunity to reduce the number of arithmetic operations performed during solution of symmetric Bayesian decision problems.

The computational efficiency of solving symmetric Bayesian decision problems with the lazy evaluation based method has been emphasized through examples.

## References

[Cooper, 1988] Cooper, G. F. (1988). A method for using belief networks as influence diagrams. In *Proc. of the 4th Conference on UAI*, pages 55–63.

[Dechter, 1996] Dechter, R. (1996). Bucket elimination: A unifying framework for probabilistic inference. In *Proc. of the 12th Conference on UAI*, pages 211–219.

[Dittmer and Jensen, 1997] Dittmer, S. L. and Jensen, F. V. (1997). Myopic value of information for influence diagrams. In *Proc. of the 13th Conference on UAI*, pages 142–149.

[Geiger et al., 1990] Geiger, D., Verma, T. S., and Pearl, J. (1990). d-separation: From theorems to algorithms. *Uncertainty in Artificial Intelligence*, 5:139–148.

[Howard and Matheson, 1984] Howard, R. A. and Matheson, J. E. (1984). Influence Diagrams. *The Principles and Applications of Decision Analysis*, Vol. II.

[Jensen et al., 1994] Jensen, F., Jensen, F. V., and Dittmer, S. L. (1994). From Influence Diagrams to Junction Trees. In *Proc. of the 10th Conference on UAI*, pages 367–373.

[Madsen and D'Ambrosio, 1999] Madsen, A. L. and D'Ambrosio, B. (1999). Independence of Causal Influence and Lazy Propagation. In *Proc. of the 5th ECSQARU*. To appear.

[Madsen and Jensen, 1998] Madsen, A. L. and Jensen, F. V. (1998). Lazy Propagation in Junction Trees. In *Proc. of the 14th Conference on UAI*, pages 362–369.

[Ndilikilikesha, 1994] Ndilikilikesha, P. (1994). Potential influence diagrams. *International Journal of Approximate Reasoning*, 11(1):251–285.

[Nielsen and Jensen, 1999] Nielsen, T. D. and Jensen, F. V. (1999). Welldefined decision scenarios. This conference.

[Shachter, 1986] Shachter, R. (1986). Evaluating influence diagrams. *Operations Research*, 34(6):871–882.

[Shachter and Peot, 1992] Shachter, R. D. and Peot, M. A. (1992). Decision making using probabilistic inference methods. In *Proc. of the 8th Conference on UAI*, pages 276–283.

[Shenoy, 1992] Shenoy, P. P. (1992). Valuation-Based Systems for Bayesian Decision Analysis. *Operations Research*, 40(3):463–484.

[Zhang, 1998] Zhang, N. L. (1998). Probabilistic inference in influence diagrams. In *Proc. of the 14th Conference on UAI*, pages 514–522.